# Frege in the Flesh: Biolinguistics and the Neural Enforcement of Syntactic Structures

Elliot Murphy

**Affiliation**: Vivian L. Smith Department of Neurosurgery, McGovern Medical School, UTHealth, Houston, TX, USA

**Correspondence**: elliot.murphy@uth.tmc.edu

# 1. Introduction

> "To speak is to translate – from a language of angels into a language of men".
>
> Johann Georg Hamann (cited in Rudd 1994: 197)

Biolinguistics is the interdisciplinary scientific study of the biological foundations, evolution, and genetic basis of human language (Givón 2002; Jenkins 2000). It treats language as an innate biological organ or faculty of the mind (Chomsky 1967; Hornstein 2024), rather than a cultural tool, and it challenges a behaviorist conception of human language acquisition as being based on stimulus-response associations (Lenneberg 1967). Extracting its most essential component, it takes seriously the idea that mathematical, algebraic models of language capture something *natural* about the world (Chomsky 2014, 2018). The syntactic structure-building operation of MERGE is thought to offer the scientific community a "real joint of nature" (Mukherji 2010: xviii), "a (new) aspect of nature" (Mukherji 2010: 235), not merely a formal artefact. This mathematical theory of language is then seen as being able to offer biologists, geneticists and neuroscientists clearer *instructions* for how to explore language (Murphy 2023).

The argument of this chapter proceeds in four steps. First, I clarify the object of inquiry for biolinguistics: not speech, communication, or generic sequence processing, but the internal computational system that generates hierarchically structured expressions. Second, I argue that this formal characterization matters for evolutionary explanation, because different conceptions of syntax imply different standards of what must be explained. Third, I suggest that a sufficiently explicit algebraic account of syntax places non-trivial constraints on candidate neural mechanisms. Finally, I consider how recent neurocomputational work begins to transform these constraints into empirically tractable hypotheses, while also noting the speculative and revisable character of the present program.

A central contention of this chapter will be the following: despite signs of its recent decline, and the ascendency of neo-connectionist models driven by the hype over Large Language Models, biolinguistics still matters because its formal characterization of syntax generates *sharper hypotheses* about neural realization than other accounts of language that prioritize processing pertaining to prediction and offer much vaguer conceptions of 'meaning integration'. This chapter will offer a number of descriptive, explanatory, and normative claims about the biolinguistic enterprise, in particular with respect to concerns of evolvability and neural implementation.

## 2. The Shape of Syntax

The term 'biolinguistics' first appeared in the title of the *Handbook of Biolinguistics* (Meader & Muyskens 1950). The authors propose that the science of language should be developed "as a natural science, and hence [should] regard language as an integrated group of biological processes", via the "functional integration of tissue and environment" (Meader & Muyskens 1950: 9). In 1974, a report of an interdisciplinary meeting on the biological basis of language again referred to 'biolinguistics' (Piattelli-Palmarini 1974), attended by Salvador Luria and Noam Chomsky, and organized by Massimo Piattelli-Palmarini, who deserves full credit for the recent popularization of the term.

Attempts to understand the biology of language bring with them "one of the most challenging sets of problems in modern science" (Fitch 2009: 283). A canonical framing begins with questions such as: What constitutes knowledge of language (the internal grammar, or 'I-language')? How is it acquired, given limited and imperfect input? How is it put to use in comprehension and production? How did this capacity evolve – and what aspects are shared with other animals, if any? A theme of the present chapter (as in much other work in biolinguistics) will be that human language is distinctly unique in the animal kingdom, in both its (putative) 'functional' properties and formal/algebraic properties. Younger apes may well make use of manual/visual signs in some manner of symbolic domain (Premack 1971), but the adult ape cannot demonstrate use of a computational-symbolic system that reaches the complexity of a four-year-old human, and many utterances of apes will be limited to requests for tickles and treats (Fitch 2009). While many humans also reserve much of their use of language for requesting tickles and treats, within the space of a couple of seconds any human child can express command of a computational resource that cyclically embeds multiple syntactic dependencies of substantial morphological complexity, going far beyond our closest ape relatives (Murphy 2016b; Murphy et al. 2022a; Rizzi 2016). Indeed, the most advanced text-to-vision AI models (Murphy, de Villiers et al. 2025) and Large Language Models (Dentella et al. 2024; Murphy, Leivada et al. 2025) also do not capture some of the most elementary components of compositional syntax-semantics. Operations in natural language pertaining to structure-dependence of rules apply to *classes of structures* (e.g., phrases), not to specific words or to linear sequences of elements (Adger 2003; Murphy & Leivada 2022).

In brief, the biolinguistic tradition has typically endorsed an *internalist* focus on I-language, rather than an externalist emphasis on speech and communication. For example, reviews that focus on the genetics of speech and communication are misleading when they equate this with language (Fisher & Vernes 2015), which is inherently modality-independent. Consider how the phonological forms of language are strongly impacted, and selected, based on external physical factors that humans are surrounded by, such as vegetation coverage, temperature, humidity, altitude, and so forth, which all impact speech properties. Yet, there is no degree of syntactic or semantic variation (at the elementary MERGE-based level) that comes about due to the density of one's immediately surrounding vegetation. The MERGE-based model of syntax common in generative grammar circles may instinctively seem too simplistic, but there are many cases in nature of simple processes and mechanisms giving rise to emerging scales of complexity and organization that would have been very difficult for to humans to predict – some of the simplest algorithms have been shown to exhibit emergent behaviors and 'side quests' (Zhang et al. 2025).

Crystallizing this architecture, Figure 1 depicts an influential conception of the language faculty within biolinguistic research circles, and the interface of language with systems for interpretation and externalization. Below are some foundational assumptions concerning the algebraic nature of MERGE, which, given the focus of this chapter, I will merely provide descriptively.

- **Commutativity** (MERGE(A,B)=MERGE(B,A)): A two-item set must ignore linear order at the moment of combination.

- **Non-associativity** ((A∘B)∘C ≠ A∘(B∘C)): Once a third element is merged, the hierarchical depth of previous combinations must be recoverable, usually through a categorial 'label' or head-selection step that privileges one member of the newly formed set.

- **Closure**: The output of MERGE is itself a syntactic object.

- **Binarity**: MERGE generates strictly binary-branching structures.
- **Non-monotonic structure-building**: MERGE can involve deletion of sub-trees or workspace elements.

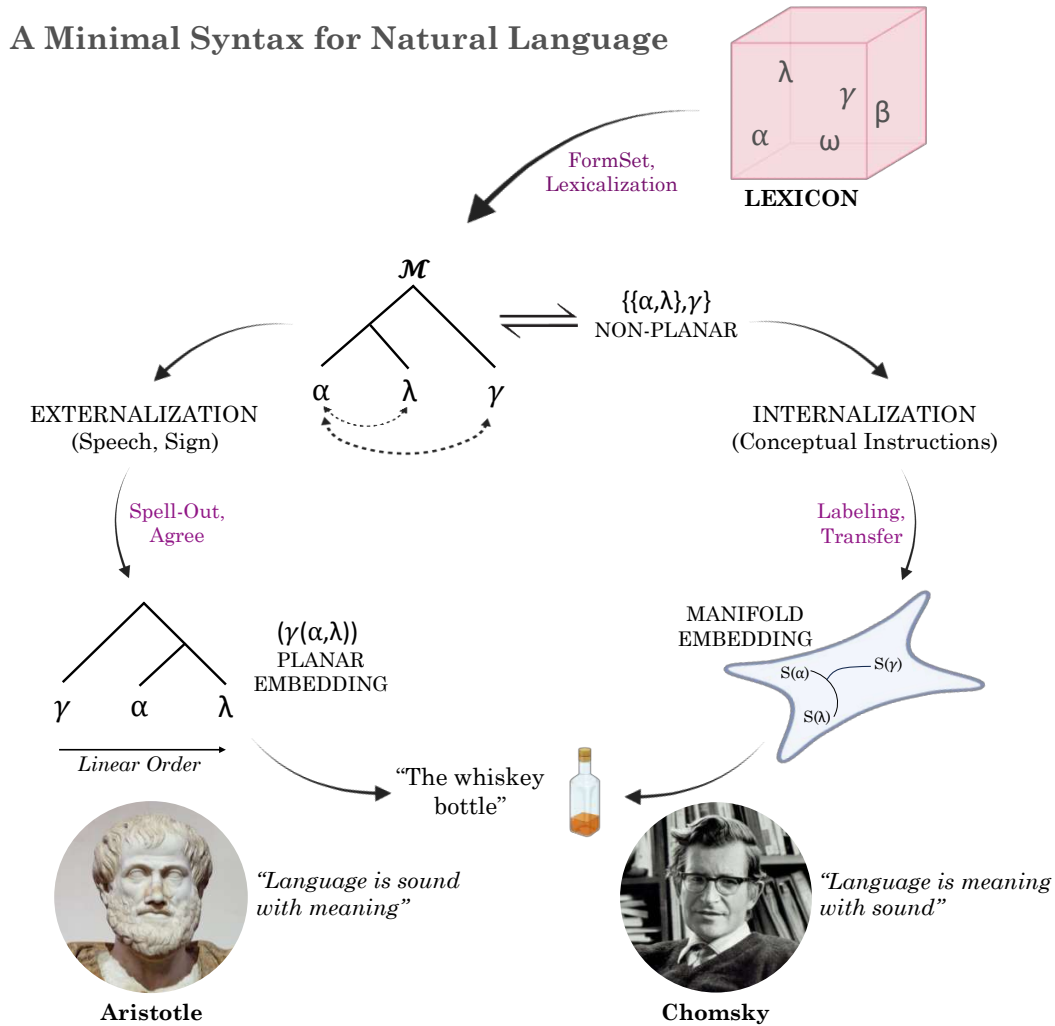

**Figure 1: Design of natural language syntax**. The architecture of the human language faculty (presented here at a higher altitude of generalization than specialist presentations in Chomsky et al. 2023; Marcolli et al. 2025; Murphy et al. 2024a). This model is strictly atemporal; MERGE is a *free non-associative commutative magma*, and generates nonplanar trees. This algebraic formulation also provides potential (though not strict) constraints on what the neural code for syntax should look like. During Internalization, structures are mapped to distinct conceptual systems (likely 'core knowledge systems'; Spelke & Kinzler 2007) with their own distinct manifold embeddings. Lexical elements ($\alpha, \lambda, \gamma$) are mapped to a workspace via MERGE ($\mathcal{M}$), which is then sent to interpretive systems and categorized ('labeled'), and/or are 'spelled-out' at externalization systems for speech, gesture, etc. For example, 'the whiskey

bottle' would be formed via {*whiskey, bottle*} being merged with {*the*}, to form a hierarchical object that is externalized in different orders depending on the language in question (i.e., not all languages adhere to the English 'Determiner Noun' order, but all human speakers converge on the same coordinates in semantic space for the phrase). The quotations at the bottom contrast opposing views on language design (see Chomsky 2011). Portrait photographs reproduced from creative commons (Wikimedia Commons; commons.wikimedia.org): Public domain photo of marble bust of Aristotle (left). Public domain photo of Chomsky from Andrea Womack (c. 1973) (right).

Biolinguistics is best understood as a convergence of (i) post-1950s formal/generative syntax, (ii) mid-20th-century neurobiology and developmental thinking about language, and (iii) late-20th/early-21st-century genetics and cognitive neuroscience, with periodic 'biolinguistics' gatherings and conferences appearing to add punctuated moments of editorial synthesis and philosophical reflection. Notable milestones, some more obvious than others, include foundational arguments in the 1950s for hierarchical syntactic structure as a worthy object of scientific explanation; the biological framing of language throughout the 1960s, with a focus on maturation and constraints on acquisition; the Principles and Parameters framework in the 1980s; and the emergence of Minimalism in the 1990s-2000s, shifting linguistics towards reducing the innate component and explaining linguistic properties via general principles of efficient computation, culminating in the MERGE-centric model of syntax. The shift from (often implicit) reliance on linear order to an order-less hierarchical structure via MERGE was also a major milestone (X-bar theory was arguably already suggestive of an abandonment of linear order within core syntax). Interwoven with these theoretical developments are milestones relating to the molecular genetics of speech and language disorders, and the cognitive neuroscience of syntax, most notably the discovery of a left-lateralized frontotemporal network for sentence processing, and various electrophysiological signatures of syntactic parsing and anomaly/repair-like processes. Lastly, often flying under the radar of biolinguistic editorial interventions, there are milestones relating to psycholinguistics, such as the development of minimalist grammars (Collins & Stabler 2016), and their use within cognitive neuroscience (Everaert et al. 2015; Friederici et al. 2017; Li & Hale 2019).

## 3. Unreal Engine: Language and the Renormalization of Thought

> "We don't have sets in our heads. So you have to know that when we develop a theory about our thinking, about our computation, internal processing and so on in terms of sets, that it's going to have to be translated into some terms that are neurologically realizable. [Y]ou talk about a generative grammar as being based on an operation of Merge that forms sets, and so on and so forth. That's something metaphorical, and the metaphor has to be spelled out someday".
>
> <div align="right">Chomsky (2012: 91)</div>

A central generative thesis, often invoked in discussions of biolinguistics, posits that learners bring a species-typical initial state to acquisition (Universal Grammar), which constrains the hypothesis space for grammars, allowing rapid acquisition from limited data. Universal Grammar is used to refer to the biological basis of whatever capacity is deemed essential and unique to language, examining the nature of 'I-language' (the internal, individual and intensional structure of language – a purely cognitive perspective) rather than 'E-language' (any other conception of language distinct from I-language, typically 'external', extra-mental conceptions of language like 'English' or 'a collection of utterances', something which could not possibly exist in the physical world) (Chomsky 2000; Leivada & Murphy 2021). The concept of I-language is, in my view, one of the most important concepts in the history of science. As Hornstein and Antony (2003: 9) summarize one of its major implications: "It is not that speakers communicate because they have an E-language in common; rather, where I-languages overlap sufficiently, communication is possible". Many operational definitions of 'biolinguistics' are tightly linked not just to the idea that this discipline studies the biology of language, but the biology of *I-language*, assumed to be amenable to naturalistic inquiry through standard scientific practices. Any individual's I-language is thought to be a token of a more general type (the faculty of language, UG, shared with other humans), not dissimilar to how Quine (1987) invokes the type-token distinction when discussing words.

In the literature, some authors when critiquing the biolinguistic enterprise have assumed (incorrectly) that any object constructed by a physical system must *also* be physical (e.g., Postal 2009), hence speaking about the biology of I-language, for these authors, is nonsensical. The claim by figures such as Postal is that language is either physical or it is Platonic, and it cannot be both. But this is a false dilemma, as Watumull (2013: 306, 308) points out:

> "The rules of arithmetic for instance are *multiply realizable*, from the analog abacus to the digital computer to the brain; *mutatis mutandis* for other functions, sets, etc. And *mutatis mutandis* for abstract objects definable as mathematical at the proper level of analysis, such as a computer program".

> "Consider how a computer program explicitly representing the Euclidean axioms encodes only a finite number of bits; it does not — indeed cannot — encode the infinite number of bits that could be derived from the postulates, but it would be obtuse to deny that such an infinity is implicit (compressed) in the explicit axioms. […] So while it is true that physically we cannot perform indefinite computation, we are endowed physically with a competence that *does* generate a set that *could* be produced by indefinite computation."

Turing's model of computation already showed why a finite physical system can encode a rule that generates an unbounded domain. Similarly, consider a computer capable of simulating a game of chess: a chess-playing program is abstract, but it is instantiated in silicon, radio waves, memory states, and so forth, across different media. Likewise, a recursive linguistic function can be abstract in kind and biologically realized in the brain. So the fact that a rule is mathematical does *not* mean it cannot be physically implemented.

Various versions of Universal Grammar have been mathematically defended mainly from poverty of the stimulus arguments, but also from some compelling arguments about the all-or-nothing format of linguistic syntax, inheriting some tools and ideas from Turing-Gödel-Post recursion-theoretic models of computability (Berwick & Chomsky 2019). In essence, the claim is that you either have the capacity for discrete infinity (recursive MERGE), or you do not (Murphy, Venkatesh et al. 2025). Existing gradualist accounts have not yet shown, in formally explicit terms, how the distinctive algebraic properties attributed to MERGE could arise through incremental recombination alone. Sympathizing with a saltationist view of language evolution, which we will return to below, Berwick (2011: 99) argues that there is zero possibility of an intermediate system between single-symbol systems, or basic concatenated sequences of single symbols, and full-fledged natural language syntax: "[O]ne either has Merge in all its generative glory, or one has effectively no combinatorial syntax at all". Others have reasonably argued that the very notion of 'proto-language' may not be defensible. Piattelli-Palmarini (2010: 160) writes: "Words are fully syntactic entities and it's illusory to pretend that we can strip them of all syntactic valence to reconstruct an aboriginal non-compositional proto-language made of words only, without syntax".

Language evolution therefore required MERGE in addition to some process of lexicalization, re-formatting certain pre-existing concepts into manipulable syntactic objects (a certain slice of conceptual space, though not all of it, for there are many conceptual representations humans have that do not lexicalize and transform into morphosyntactic objects). Language in the 'narrow' sense is just this: MERGE and its objects. Then, when this system is connected to relevant conceptual interfaces and sensorimotor interfaces for externalization, we get the full list of regular phenomena that most linguists typically think of as language, e.g., morphosyntactic and morphophonological representations such as word order differences across languages, agreement, island constraints, case marking, *wh*-movement, *wh*-in situ, head movement, scope and interpretive asymmetries, predicate-argument structure, and other forms of displacement.

For these reasons, the thesis that human-unique intellectual capacities emerged purely from an expanded capacity for information processing (Cantlon & Piantadosi 2024) cannot be the full story. For example, in Cantlon and Piantadosi's (2024) discussion of relational reasoning, they acknowledge that symbol training dramatically helps non-human primates succeed at quaternary tasks. But this is actually a problem for a pure capacity story: if it were just about holding more items in working memory, why would an externally-provided symbol unlock performance so dramatically? The symbol seems to be doing representational work (compressing or reformatting the problem) not just offloading memory.

Still, there is something to this intuition from Cantlon and Piantadosi (2024), and it is similar to Rizzi's (2016) ideas about different Merge-based systems (i.e., word-word, word-phrase, phrase-phrase systems). Morever, Chen, Brincat et al. (2026) show that constructing hierarchical structures optimizes the processing efficiency of sequential language input while staying within memory constraints, potentially contributing to an explanation for at least the general structure that hierarchical syntax takes.

Within biolinguistics it has been common for researchers to assume a tighter connection between the format of certain types of human thoughts and the mechanisms of natural language syntax-semantics, whereby language does not merely passively *express* a set of pre-existing (and pre-linguistic) conceptual structures, but also serves to re-format types of conceptual configurations via lexicalization, reference, predication, and truth-conditional content (Davidson 2004; Hinzen 2014; Huybregts & Riemsdijk 1982; Lenneberg 1967; Murphy 2017; Pietroski 2018; Pustejovsky & Batiukova 2019; Schein 2017). MERGE-based syntax permits a kind of information coarse-graining, re-formatting concepts and giving us new kinds of 'things' to think about. The evolution of language, within the biolinguistic tradition, is thought to have "offered humans ways of planning, critically assessing, engaging in fiction and fantasy, and inquiring that are unavailable to other creatures" (McGilvray 2017: 6). These forms of meaning are unlike anything else in the animal kingdom, as Hinzen (2014: 231) notes:

> "[C]onceptualization in non-linguistic beings is still continuous with perception; it also remains stimulus-controlled, non-combinatorial, and non-propositional, and concepts are not employed for purposes of intentional reference, with a capability to refer to anything at all no matter how remote in space and time".

A recent editor's summary of a *Science* paper by Arnon et al. (2025) argues that "recent research has made clear that language per se is not unique to humans". The term "per se" is doing a substantial amount of work here: while Arnon and colleagues successfully show that elements of the broad language faculty are shared with non-humans, there is no demonstration that the core faculty of language (unbounded, recursive MERGE) is shared. For example, it is unclear how one might go about "culturally transmitting" the discrete mathematical capacity for implementing a free, non-associative, commutative magma operator in an unbounded manner. A major problem with this line of research into iterated cultural learning is that these studies subtly presuppose that some Merge-type combinatorial operator is antecedently present prior to "cultural transmission", which can certainly help us uncover interesting properties about the development of morphophonological and morphosyntactic complexity over time, but it still begs the question as to why and how language evolved. Of course, this is not to say that the abrupt emergence of MERGE should not be seen as part of a larger biocultural architecture for how the broad faculty of language evolved. This perspective can therefore help furnish more of a *tempered and scaffolded saltationist* view, as argued elsewhere (Murphy 2019).

In this view, many 'semantic networks' are thereby defined as what kinds of meanings we can generate in the *absence* of grammar (e.g., knowledge of individual lexical items like *canoe*, or the associations its features have with other lexical items, or the nodal and hierarchical structure we can derive from *canoe* in terms of it being a type of boat, which is in turn an inanimate entity). Once grammatical MERGE-based systems are entertained, new forms of inference can be derived, and we can refer to an unbounded array of possible situations, events, substances, facts, and objects, and propositions about these might or might not be true. The meaning of *I wished our king wasn't our king* does not come from the meaning of individual words alone (the determiner phrase *our king*, at the D+NP level, is the same twice), but from the grammatical configuration of subject-predicate. As the early Wittgenstein suggested, we might think of the world as the totality of facts, and if grammar permits us to make truthful statements then grammar gives us a novel conception of 'the world', even if it does not *exhaust* our notion of the world. Indeed, as grammatical and morpho-syntactic complexity increases, so too does referential definiteness (i.e., I ate dog < I ate dogs < I ate a dog < I ate the dog < I ate this dog) (Hinzen 2014), a relation that does not seem exclusively anchored around lexico-semantic content, but also grammatical configuration. Hence grammar affords the generation of more precise coordinates in conceptual space; in the case of the 'dog' sentences, it fixes whether we think of a mass, a set of non-specific individuals, one specific individual, one part of an individual, and so forth. If human thought was exhausted by 'semantic associations' alone, then we would have no such things as predicate-argument relations, subjects, topics, presuppositions, entailments, or truth-values (or 'truth-indications', for internalists). To put this another way: Purely associative systems encode similarity, but linguistic thought requires structured relations.

Moreover, it is not quite true to say that grammar allows us to "combine concepts", as is commonly said: it is more that concepts are placed into a grammaticalized form and then *those grammatical elements* (parts of speech, phrases) are combined. Even in the case of compounds ('dog food'), grammar never *combines concepts*, but rather a head is combined with a modifier, which are linguistic elements that later provide instructions to conceptual spaces. Interesting evidence that part-of-speech distinctions are not lexical but are in fact already subserving grammatical functions comes from the observation that grammar can overrule lexical part-of-speech specifications (e.g., "topping the agenda"). These observations are commonly entertained within biolinguistics, but they depart quite substantially from what was believed from Frege to Russell, and from Carnap to Quine; namely, that language was effectively a poor translation of logical form.

Per the Church/Turing thesis, computation is typically discussed in theoretical linguistics as "just recursion defined over a finite set of primitive functions" (Collins 2008: 49). Yet a commitment to a certain number of innate factors does not entail commitment to some form of gene-centrism, since the environment clearly also plays an important role. Language and cognition are therefore thought to be parts of the natural world on a par with *chemical*, *optical* and *electrical* aspects, being just as amenable to naturalistic inquiry (Ueda 2016). As Epstein et al. (2017: 51) make perfectly clear, "vision scientists are not directly engaged in trying to account for *what* people might decide to look at"; rather, they are interested in explaining how we use our eyes to see in the first place.

Following Hauser and Watumull (2017), many researchers aligned with the biolinguistic enterprise invoke a universal governing generative system to subsume language, arithmetical thought, musical thought, artistic thought, and even moral thought. This Universal Generative Faculty (UGF) produces a mental architecture through which distinct representational domains access a universal Turing machine which provides a number of generative procedures, and which differ based on which representational domain interfaces with UGF: recursion for language and music (McCarty et al. 2023), iteration for motor routines, and the successor function for mathematics. As the Hauser and Watumull (2017) write:

> "*Computability* is common to all generative systems, and *definition by induction* is ubiquitous in the generation of hierarchical structure. Thus it is possible for a system to be computable but not generative of hierarchy (e.g., Lashley's motor routines), as well as computable and generative of hierarchy but not unbounded (e.g., birdsong or human phonology)."

Different representational domains make use of UGF in distinct ways. For example, morality seems to access concepts which are only derived from *content* words like nouns and verbs, with most moral judgements involving certain actors doing things to others. *Functional* elements, like determiners, auxiliaries and copulas, seem language-specific, and indeed the combination of content and functional elements seems to be the core representational feature of human language syntax (Miyagawa et al. 2013). Indeed, while function words are very often derived from content words, content words are virtually never descended from function words, suggesting a clear evolutionary trajectory for the complexity and organisation of language (Hurford 2014). This is an ambitious and in some respects speculative proposal: for example, the evidence that moral cognition and musical cognition exploit the same underlying recursive operation as syntactic computation is presently thin, and the differences in representational format across these domains are substantial.

Nevertheless, some type of shared generative capacity across domains seems to derive many species-specific components of cognition (Dehaene et al. 2022), producing a range of novel (though often ill-defined) representations (Favela & Machery 2023). Descartes's *Compendium on musick*, published posthumously, claims music to be a "passion of the soul" and examines some of its mathematical foundations and how formal structure determines aesthetics. These sorts of instincts drive much of the work conducted under the umbrella of biolinguistics – exploring how much of nature can be carved out by only invoking simple mathematical rules and procedures. Biolinguistics is almost intrinsically committed to the representational theory of mind, or the notion that in order to understand cognition we must posit the existence of internal cognitive states that serve to represent fixed entities, properties, and events (Fodor 1975; Neander 2017). Alongside a seemingly species-specific syntactic capacity, humans also display remarkable *representational* capacities (Krakauer & Ramsey 2026) – it remains an open question to what extent these two domains are related.

Horwich (2003: 165) has described the internalist (or mentalist) theory of language in this lucid and concise summary, in which he details how the biolinguistic program assumes the following:

> "(a) that each human being indeed has a faculty of language, FL, a component of his mind-brain constituting the primary causal/explanatory basis of his linguistic activity; (b) that the possible states, L1, L2, [...] , of FL are, by definition, possible I-languages; (c) that each such state, L, is a computational procedure that generates infinitely many I-expressions, E1, E2, [...]; (d) that each such expression, E, is a pairing <PHON(E),

SEM(E)> of phonetic and semantic objects, which, through their respective interaction with the perceptual/articulatory system (P/A) and the conceptual/intentional system (C/I), determine an association of a sound with a thought; (e) that these PHON-SEM pairs are constructed from lexical items, LI1, LI2, [...]; and (f) that these lexical items are stored in a lexicon which is accessed by the computational procedures that form I-expressions".

Before we progress further into the biology of language, I would be remiss to ignore the following series of questions that have often been discussed in relation to the core goals of biolinguistics (Chomsky 1986, 1988):

1. What is knowledge of language?
2. How is that knowledge acquired?
3. How is that knowledge put to use?
4. How is that knowledge implemented in the brain?
5. How did that knowledge emerge in the species?

Due to the sheer scope of these questions, the remainder of the present chapter will not aim to provide each of these with equal treatment. Indeed, given that the focus of this volume is syntax, I will naturally restrict the present discussion to how the biolinguistic enterprise approaches this topic, as opposed to pragmatics, second language acquisition, phonology, and other areas of the language sciences. More specifically, I will ultimately focus not just on the neural 'implementation' of language, but on a more precise question of neural realization that can now be asked, thanks to the past decade of research: Which neural processes enforce the algebraic properties of natural language syntax? By *neural enforcement* I mean the causal realization of formal constraints during online parsing, and the functional preservation of relevant distinctions about the algebraic properties of language at particular scales of neural organization. Neural enforcement does not mean that the brain literally contains sets, trees, or category-theoretic objects in explicit symbolic form. Nor do I assume a one-to-one mapping between formal objects and localized neural states. The weaker and more defensible claim is that if syntax has distinctive algebraic properties, then successful neural theories must preserve those distinctions at some implementational grain, whether through oscillatory coordination, recurrent circuit motifs, state-space geometry, or other mechanisms not yet adequately described.

## 4. Plato vs Darwin: Laws of Form and Natural Selection

The biolinguistic enterprise has typically advocated for a saltationist account of cognitive evolution (a punctuated, sudden and large macro-mutational change across generations), followed by a gradualist phase interfacing MERGE-based syntax with other representational systems. Some comprehensive evolutionary timelines have been presented in the literature concerning when MERGE, the lexicon, and the interfaces may have emerged (Arnon et al. 2025; Chomsky 2010, 2012; Markov et al. 2023; Murphy 2012, 2019; Murphy & Benítez-Burraco 2018a). An adaptationist perspective argues that language (including grammar) can be explained as being shaped by natural selection for communicative or cognitive functions, with design-like features reflecting functional pressures. A contrasting family of views, more commonly cited within biolinguistics, emphasises exaptation (co-option): features can acquire current utility without having been built by selection for their current role (Gould & Vrba 1982). The conceptual distinction between adaptation and exaptation was argued to be essential for evolutionary explanation in general, and it is often invoked in language evolution debates when suggesting that some aspects of syntax may be by-products or co-opted capacities rather than direct adaptations for communication (Murphy 2020b). Some biolinguistic proposals combine these perspectives by reserving a narrow syntactic core as possibly minimal and recently emerged, while treating many communicative and externalization aspects as exploiting older systems. Other approaches claim that potentially most, if not all components of language are shared with other species, but it is the unique combination in humans that gives rise to species-specific intellectual capabilities (Arnon et al. 2025).

An implicit assumption in many biolinguistics publications is the idea that interdisciplinary boundaries are a nuisance, artificial, mostly sociologically imposed, and that the language sciences need to embrace trans-disciplinary perspectives. This mirrors closely the methodological naturalism of many stands of the generative grammar enterprise (see the papers collected in Anthony & Hornstein 2003).

Within this tradition of biolinguistics, syntax is often taken to be central, because a finite system appears to generate an unbounded range of structured expressions, and because many hallmark properties of human language (hierarchical constituency, long-distance dependencies, compositional interpretation) track this generative capacity (Murphy & Woolnough 2024). Biolinguistic treatments of other aspects of language are not difficult to find (Di Sciullo & Boeckx 2011; Samuels 2011), but syntax and the means by which it regulates semantic instructions very much remain the central focus (for reasons that can be traced back to Chomsky 1957, 1963, 1965, 1968).

Though they are by no means synonymous, a major strand of biolinguistic work aligns with generative grammar, especially the Minimalist Program, where a small number of operations (notably, the *selection* of lexical elements and MERGE) build hierarchical syntactic objects that interface with conceptual-intentional systems and sensory-motor systems (Chomsky 1995). This strand frames explanation in terms of (i) an innate endowment (Universal Grammar), (ii) experience/data, and (iii) domain-general 'third-factor' principles such as computational efficiency, laws of form, and general constraints on biological systems (Chomsky 2005).

More broadly, the biology of syntax has been approached through converging methods: behavioral and molecular genetics (candidate genes, sequencing; Poeppel 2011), neuroimaging and electrophysiology (fMRI, MEG, intracranial EEG; Murphy et al. 2022a, 2022b, 2023, 2024b, 2026), causal neuropsychology (lesion-symptom mapping; Hickok 2025; Murphy 2026; Zaccarella & Trettenbrein 2021), developmental and cross-linguistic acquisition studies (including emergent sign languages; Trettenbrein et al. 2021), comparative cognition (animal learning; Zuberbühler 2019), and computational modelling (learnability theory, iterated learning, evolutionary dynamics; Bunyan et al. 2025; Leivada & Murphy 2022).

Although Darwin successfully managed to bridge many of the classical gulfs separating humans and other animals, many of his contemporaries, such as the linguist Müller (1866), complained that "language is the Rubicon, which divides man from beast, and no animal will ever cross it". A prominent hypothesis differentiates language capacities that are broadly shared (e.g., perception, vocal learning components, general memory) from a narrower, possibly uniquely human computational core, often termed the narrow faculty of language (Hauser et al. 2002, 2014). Even if one does not subscribe to the biolinguistic focus on syntax, it is difficult to deny how methodologically productive the study of syntax has been when one recognizes how it has been used as a kind of bridge discipline: Syntactic theory provides explicit hypotheses about representations and operations (e.g., constituency, agreement, movement), which can be tested against developmental trajectories, neurological dissociations, genetic syndromes, and comparative cognition tasks in ways that are harder for purely descriptive approaches (for discussion, see Friederici 2011, 2012, 2016, 2017; Pylkkänen 2019). Multiple lines of work support a distinction between hierarchical composition (often treated as core) and linear order (often treated as part of externalisation). Evidence that the language-selective network in the brain remains strongly engaged under word-order degradation (so long as local composition is still plausible) has been interpreted as supporting the primacy of composition over surface order in neural responses (Mollica et al. 2020), but it also aligns with theoretical views that treat MERGE-built structures as fundamentally order-free and view ordering as a separate mapping problem.

Meanwhile, child language studies have been used to argue that learners prefer structure-dependent generalisations rather than linear-order heuristics, consistent with innateness claims about hierarchical bias (Pearl 2022; Perkins & Lidz 2021). Experiments with English-speaking children indicate that 'children are genetically predisposed to rule out structure-independent options' when forming *yes/no* questions involving auxiliary inversion (Sugisaki 2016: 126). Language differs from animal systems of communication in that it is not stimulus-driven, nor does it have a specialised function like alarm calls or seduction songs.

Biolinguistics does not typically dictate a single syntactic framework, but it does impose unique *pressures* on syntactic theorising – especially on what counts as a good explanation. With respect to Minimalist criteria, it is typical within biolinguistics not to simply assume that language is perfect or optimal in some absolute sense, but rather something closer to the following: Given interface conditions and third-factor constraints, *some* aspects of syntactic computation may exhibit surprisingly simple design. The philosophy underlying most machine learning approaches, or in the neurolinguistic space brain-LLM alignment scores, are not the primary goal of linguistic research, but rather what Chomsky (2022) terms 'genuine explanation'. When syntax is treated as a biological capacity, syntactic theories are pushed to specify computational primitives (operations like MERGE, labeling, feature checking/unification, constraints on locality) (Adger 2013, 2022; Murphy 2015a, 2015b; Murphy & Shim 2020), interfaces to meaning and externalisation, and developmental implementability. Minimalism is explicitly shaped by this pressure, aiming to reduce stipulative machinery and to derive properties from general principles and interface conditions. Relatedly, although it is often said that "[y]our theory of language evolution depends on your theory of language" (Jackendoff 2010) – and Jackendoff is surely right that this is *often* the case – there are many theories of language that are presented as gradualism-compatible that are *also* compatible with saltationist models of evolution (e.g., Jackendoff's own parallel architecture is fully compatible with a saltationist account).

Although language evolution has typically occupied centre stage, biolinguists concern themselves with a wide number of different research questions, employing frameworks general to evolutionary theory. As mentioned already, whichever research topic is pursued, there are by necessity three independent factors which interact to generate the final state of any biological property related to language (Chomsky 2005): (1) genetic endowment, (2) environment ('stimuli', 'data') and (3) principles not specific to language (general physical law). Importantly, much of the relevant 'environmental' information in epigenetic interaction is the local environment surrounding each cell, and not the external world. Meanwhile, the general physical laws in (3) impact the efficient wiring of the brain (e.g., dynamical system constraints) and electrophysiological dependencies between brain rhythms and states. Fermat's principle of least time seems to apply to how light travels and also to how insects such as fire ants navigate space (via the shortest path) (Oettler et al. 2013). These principles of efficiency are based on natural law, not natural selection. Another possible example of a third factor is the Input Generalization by Holmberg and Roberts (2014), the learning strategy which generalises from detection of one instance of a category to all future instances; for example, after seeing a black bat, assume all bats are black until contrary evidence presents itself. Relatedly, human-like language learning may arise from strong, structured inductive biases operating within flexible neural systems (McCoy & Griffiths 2025). At least one clear type of learning bias that humans seem to have, which is typically embraced by both generative and non-generative linguists, and which is either absent or much weaker in other species, is a domain-general proclivity to infer tree structures from data whenever possible – sometimes termed *dendrophilia* (Fitch 2014).

Turning now towards strands in philosophy of biology that have attracted researchers within the biolinguistic enterprise, we can contrast two previously mentioned approaches. Adaptationist accounts and exaptation/constraint-based accounts differ not only in conclusions but in standards of explanation. Adaptationist accounts foreground function and selection; exaptation accounts foreground co-option, architectural constraints, and third-factor explanations. The neo-Darwinian synthesis (closely related to functionalist and adaptationist models of evolution, which use a biological system's putative function to direct enquiry into its origins) has great power with regard to explaining survival, and provides some insights when investigating the form and structure of organic systems, but – so the biolinguistics story goes – it does not provide full explanations of language design. Indeed, functionalism and the neo-Darwinian synthesis, by focusing on function rather than form, face certain difficulties even with supposedly simple structures like bones, which simultaneously provide bodily support but also store marrow and calcium for producing red blood cells, and so could be regarded as part of the circulatory system.

Gradualist philosophies have driven a substantial amount of work on language evolution. Notably, Pinker and Bloom (1990: 713) assume that "language is a complex system of many parts, each tailored to mapping a characteristic kind of semantic or pragmatic function onto a characteristic kind of symbol sequence". For Pinker and Bloom, discrete building blocks include lexical categories, phrasal categories, phrase structure rules, rules of linear order, mechanisms of control, *wh*-movement, amongst other things – each driven by small evolutionary steps. In contrast, Chomsky has long argued (e.g., first via the reduction of complex phrase structure and movement rules to External/Internal Merge (Chomsky 2004a), and then the reduction of both of these to a single MERGE computation) that much of these superficial complexities can be derived from a simpler underlying system of syntax-semantics and their interfaces (Chomsky et al. 2023; Drummond & Hornstein 2011), hence the need for many punctuated moments of evolutionary innovations is rendered less acute.

As Zeder (2017) summarises: "In the [modern neo-Darwinian synthesis] natural selection is recognized as the preeminent and ultimate causal force in evolution that sorts variation arising through random mutation, and passes on adaptive variations at a higher rate than less adaptive ones, resulting in an evolutionary process that proceeds at a gradual pace made up of small microevolutionary changes in the composition of individual genes and alleles within genes". Fuentes (2016) pushes for an 'Extended Evolutionary Synthesis', which takes into account the full range of biological, cultural and psychological factors underpinning the evolution of the human condition. Certain strands of evolutionary biology have moved beyond the 'micro mutational' gradualist perspective best symbolised by the hill-climbing metaphor, that was derided by William James as 'Pop-Darwinism' in his critique of Herbert Spencer, but which is still often invoked directly in rebuttals of saltationism (e.g., Kinsella & Marcus 2009). Separate research programmes generate separate research methods, and in evolutionary psychology (aligned to a large extent with functionalism) assuming a priori that a given trait is an adaptation is an experimental heuristic, an assumption not always made in other strands of evolutionary theory. Adaptationism assumes that language has evolved to fulfil a particular function given some form of pressure/requirement. But what are often viewed as adaptive problems are also not necessarily so, but could also, on occasion, be the "regularities of the physical, chemical, developmental, ecological, demographic, social, and informational environments encountered by ancestral populations during the course of a species' or population's evolution' (Tooby & Cosmides 1992: 62). Adaptationism at times seems to hearken back to August Weismann's (1893) argument for the *Allmacht*, or omnipotence, of natural selection over other evolutionary forces.

Under a non-adaptationist view, language may constitute an exaptation, or alternatively a spandrel, a feature not directly selected for but a by-product of evolution (Piattelli-Palmarini 1989). The computational mechanism underlying language may have arisen as a by-product of general principles of neural organization. Once present, however, this capacity could have been exapted for communication and symbolic thought. More specifically, and moving somewhat beyond the basic exaptation hypothesis, language might reflect constraints from mathematical or morphogenetic laws of form, meaning that the combinatorial algebra of syntax emerges because biological systems naturally implement certain classes of computation. This will remain a theme throughout the rest of this chapter. This emphasis on using mathematical tools to fundamentally constrain and guide scientific inquiry across all organizational scales is similar in spirit to Plato's belief that intelligibility was to be found only in the world of geometry and mathematics, with the complex world of sensation being unapproachable. An effective study of astronomy, in his view, requires that "we shall proceed, as we do in geometry, by means of problems, and leave the starry heavens alone" (1945: 248–249).

Functionalist and adaptationist models will certainly be useful in exploring certain aspects of (broad) language evolution (for instance, imitation and speech perception), but it is typical within the biolinguistics tradition to assume that core compositional operations for syntax-semantics are less amenable to this treatment. The competition between functionalism and 'formalism' (to use nineteenth-century terminology, focusing on physical form and structural complexity independent of environment) will be ever-present in discussions of biolinguistics, as it is in much of the broader biological sciences. For example, discussing the physical genesis of multicellular forms, Newman et al. (2006: 290) conclude that, "rather than being the result of evolutionary adaptation, much morphological plasticity reflects the influence of external physicochemical parameters on any material system and is therefore an inherent, inevitable property of organisms". While functionalism is concerned with structural use, formalism stresses the importance of physical principles determining organic structural complexity. Whereas many have characterised the central debate in nineteenth-century biology as being between evolutionists and creationists, a more accurate classification (as Darwin himself noted) would distinguish teleologists (who regarded adaptation as the single most important aspect in evolution) with morphologists (who held that commonalities of structure were the defining biological characteristic) – a dichotomy stressed by E.S. Russell's *Form and Function* (1916). The Modern Synthesis (Neo-Darwinism) can often be fairly weak at explaining *why* biological forms have the specific structural properties they do, beyond saying they were selected.

Some other formalists include such figures as D'Arcy Thompson, Brian Goodwin, Richard Owen, Stuart Kauffman, Geoffroy St. Hilaire, Richard Goldschmidt, Nikolai Severtzov, Louis Agassiz, Karl Ernst von Baer, and Goethe (whose plant studies led him to coin the term "rational morphology", and who posited what he called the "Urform", a class of physically possible organisms, a concept familiar to many contemporary developmental biologists such as Michael Levin). They focused on form and structural commonalities as their explanadum, leaving aside the question of adaptive effects as a secondary concern. Richard Owen, for instance, stressed what he called "Unity of Type" over "conditions of existence" – concerns which theoretical biologists like Thompson would later emphasize. Darwin's rival Wallace also objected to the claim that natural selection alone could explain human arithmetic capacities, scarcely employed throughout history and largely hidden in the head. Darwin (2007: 84) himself proposed in *The Descent of Man* that the only distinction to be made between humans and other animals was that man differs "solely in his almost infinitely larger power of associating together the most diversified sounds and ideas; and this obviously depends on the high development of his mental powers". To adopt the essence of Wallace's point, 'evolution' involves many factors, not just natural selection.

These internalist concerns are also found in the perceptive words of the seventeenth-century German physician Daniel Sennert (1650: 765):

> "Always in vain does anyone resort to external causes for the concreation of things; rather does it concern the internal disposition of the matter [...] On account of their forms [...] things have dispositions to act [...] They receive their perfection from their form, not from an external cause. Hence also salt has a natural concreation [...] not from heat or cold, but from its form, which is the architecture of its domicile".

Reviving these considerations, biolinguistic internalism is opposed to a Lamarckian interpretation of evolution in which external factors determine internal structure, with internal structure satisfying these external demands. Adaptationism is not to be discouraged altogether, and can contribute to evolutionary theory so long as it arises where laws of form have constrained the route ahead. As Hinzen (2006: 15) rightly notes, the existence of something does not necessarily concern the *nature* of that thing, a notion which "surfaces in Sartre's famous existentialist argument that there is no such thing as human nature, for from mere *existence* (Heidegger's *Geworfenheit*), no human *nature* could be extracted. The existentialist's point is that the conditions under which a thing exists do not only leave its nature *underspecified*; as such they do not determine it *at all*".

The goals of theoretical morphology, outlined by McGhee (1998: 2), are in some ways the closest to those of biolinguistics:

> "The goal is to explore the possible range of morphologic variability that nature could produce by constructing n-dimensional geometric hyperspaces (termed "theoretical morphospaces"), which can be produced by systematically varying the parameter values of a geometric model of form. [...] Once constructed, the range of existent variability in form may be examined in this hypothetical morphospace, both to quantify the range of existent form and to reveal nonexistent organic form. That is, to reveal morphologies that theoretically could exist [...] but that never have been produced in the process of organic evolution on the planet Earth. The ultimate goal of this area of research is to understand why existent form actually exists and why nonexistent form does not".

The intuition behind these proposals has a long history, as indicated already. L.T. Hobhouse remarked in his *Mind in Evolution* that the chaotic motion both of long grass and of "the white blood-corpuscle" are "only very complicated results of the same set of physical laws in accordance with which the grass bows before the wind" (1901: 11–12). Hinzen's (2006: x) position, and that of many within biolinguistics more generally, is "essentially parallel to one found in theoretical biology, where a position that its nineteenth century defenders called 'formalism' or 'rational morphology' allowed for the autonomous study of animal form, disregarding the external *conditions of existence* that drive such organic forms in or out of existence on the evolutionary scene". The final shape that the human brain takes over development seems clearly to be constrained by both genetics and physics (Van Essen 1997, 2023; Xu et al. 2010). For more compelling examples of physical constraints helping to explain organic forms, see Newman and Bhat (2009) and Turner et al. (2020).

In addition, the brain's learning mechanisms are constrained by dendritic compartmentalization and vectorized signaling, with cortical dendrites carrying vectorized instructive signals supporting credit assignment (Francioni et al. 2026); an example of third-factor constraints on neural computation. Other recent work indicates that brain dynamics are fundamentally constrained by geometry (shape), not just connectivity. The brain cannot implement arbitrary dynamics, and is restricted to eigenmodes imposed by its geometry. One of the clearest modern demonstrations of a third-factor constraint on neural computation comes from Pang et al. (2023). These authors show that large-scale neural dynamics are constrained by the geometry of the cortical manifold, which imposes a fixed eigenmode basis on brain activity. Within this framework, oscillatory mechanisms such as phase-amplitude coupling can be understood as operating over geometrically defined modes (Murphy 2025c).

Luo's (2021) review of neuronal circuit architectures is also useful for biolinguistics because it shifts explanation away from isolated neurons and toward reusable motifs and larger-scale connectivity plans that implement specialized computations. In this framework, microcircuit motifs function like "words" and larger architectural plans like "sentences". Although Luo does not discuss language specifically, his central claim – that cognition depends on evolutionarily and developmentally constrained circuit architectures built from conserved motifs – fits well with generative approaches that treat language as a structured computational system rather than a mere statistical association space.

What these diverse findings share is a common implication: the space of possible neural computations is not unconstrained. Cortical tissue is not an infinitely plastic substrate that can implement any arbitrary algorithm; rather, it is a structured medium whose geometry, connectivity motifs, and dendritic architecture pre-select certain classes of computation as natural and others as difficult or impossible (Makin & Krakauer 2023). For biolinguistics, this means that the question 'how does the brain implement MERGE?' is not answered by finding a region that 'does syntax', but by identifying which of these pre-selected computational classes is compatible with the algebraic properties of syntactic structure.

Lewontin's 'Triple Helix' (of environment, genes and the organism) (Lewontin 2002), Gould's (2002) 'historical', 'functional' and 'formal' causal influences on the creation of natural objects, and Chomsky's 'three factors' in language design (data, genetic endowment, natural law) lead to a familiar approach in biology: examine organism-internal constraints, with any analysis of environmental and 'cultural' events taking place only within these constraints. Koonin's (2009: 1011) summary is instructive here:

> "Evolutionary-genomic studies show that natural selection is only one of the forces that shape genome evolution and is not quantitatively dominant, whereas non-adaptive processes are much more prominent than previously suspected".

Some authors (such as Coyne 2010, but he is by no means alone) even go so far as to explicitly note that they will interchangeably use 'evolution' and 'natural selection', as if these concepts are equivalent. The popular focus on natural selection is of course justified to some extent, since positive selection is well known to have acted on the human genome (Sabeti et al. 2006), and brain areas putatively advantageous or crucial for language processing (such as the frontal lobe, and Broca's and Wernicke's areas) are under substantial genetic control and are highly heritable (Thompson et al.

2001). But focusing on natural selection should not lead to the exclusion of other evolutionary forces. Many contemporary philosophers and intellectuals propose a stark choice between God and natural selection (e.g., Dennett 2018: 277), as if those are the only two options, which somewhat misses the point for the purposes of constructing accurate evolutionary theories, since the explanatory space between divine creation and natural selection is vast and scientifically productive (Jenkins 2000). As Hinzen (2006: 96) points out, the functionalist doctrine articulated by Pinker ("[H]ow well something works plays a causal role in how it came to be" (1999: 162)) "flies in the face of the Darwinian doctrine that neither adaptation nor natural selection are sources of genetic change. Mutations are the sources of novelty, their causes are internal, and they are crucially undirected, hence not intrinsically functional".

Figures such as Richard Dawkins, Daniel Dennett and Steven Pinker, amongst other Neo-Darwinian synthesis scholars, have in various places argued that the principle of Darwinian evolution is the only possible way that something can improve by itself – nothing in the world gets more adaptive except by this mechanism of natural selection. The mechanism of selection chooses from an existing set of things a subset to retain, and discards the rest; it starts from a population of things that are distinct from each other, and self-reproducing. Others have argued that natural selection is not the only possible mechanism of creative adaptation.

This is not to dismiss the adaptationist research program wholesale. It has generated productive, specific predictions about the peripheral systems of language (vocal tract anatomy, auditory processing, social learning, and imitation) and has motivated significant comparative work on communication in other species. But disagreement lies in whether natural selection for communicative utility is the right explanatory lever for the core combinatorial properties of syntax, or whether those properties reflect constraints that lie upstream of selection pressure.

In contrast, Watson, Levin and Lewens (2025) have argued that *natural induction* is a strong candidate to work alongside the logic of selection. This is a recent biological framework that resonates with third-factor and law-of-form thinking by challenging the monopoly of selection-based explanation (Buckley et al. 2024). Natural induction is a process that starts from a population of things which are all interacting, specifically in dynamical systems, and the nature of the *relationships* between things are changed, rather than discrete things being discarded. As those relationships change, the organization of the system as a whole is changed, potentially granting novel computational affordances that would otherwise be dormant. While for natural selection the *deductive* nature is foregrounded, and any inductive component is implicit, with natural induction this is reversed. Unsurprisingly, evolution by natural induction is highly amenable to being applied to complex systems, such as brains. The core claim of Watson et al. (2025) is that via natural induction dynamical systems with interacting parts, recurring perturbation, and slightly plastic/viscoelastic couplings can reorganize themselves in a learning-like way, storing past constraints and generalizing from them. They argue that this can produce adaptive organization without natural selection and that, in some evolutionary scenarios, natural selection functions more as a memory or canalizer of solutions found first by natural induction than as the source of those solutions. That renders it highly resonant with third-factor thinking in biolinguistics, whereby language design is not to be explained solely by direct adaptation for communication; one must also consider principles not specific to language and sometimes not specific to biology at all.

Like many within the biolinguistics tradition, Watson et al. (2025) argue that adaptive structure can emerge because systems obey general learning-like and morphogenetic principles, and selection may merely stabilize or inherit what those principles already discover. Watson et al. (2025) state that, "[s]ince induction is not about differential survival or reproductive fitness, the adaptations it explains may be different in kind". A standard adaptationist view gives explanatory primacy to selection acting on heritable variation. Watson et al. explicitly reject the idea that selection has an exclusive explanatory monopoly. They argue that adaptive phenotypic plasticity can arise first, and then genetic evolution can assimilate it afterwards, such that selection does not 'discover' the adaptive phenotype but only retains or canalizes it. As Watson et al. (2025) put it: "if natural selection is not the sole mechanism of evolutionary adaptation, a lot of evolutionary biology may need a rethink".

Following this revised evolutionary logic, language-related biological organization may have been shaped not only by selection for internal cognitive enhancements via MERGE-based structures, and later communicative success, but also by generic constraints on how developmental, neural, bioelectric, and regulatory systems self-organize, learning-like dynamics in morphogenesis and multiscale physiology, and phenotype-first processes in which certain structured capacities appear because the organismic system naturally settles into especially stable, compressible, or computationally tractable forms. Researchers should therefore consider a biolinguistic stance that says: before asking what selection optimized language for, ask what kinds of structured systems are easiest for living matter to build under general laws of organization.

This allows us to link the standard Chomskyan (2005) perspective (that language design may reflect general principles of computation, development, and physical law, not just selection), with ideas about how adaptive organization in biology may come from general learning-like dynamical principles, with selection often acting downstream as a stabilizer rather than an inventor, in addition to related views about morphogenesis from figures ranging from Alan Turing and D'Arcy Thomson to Michael Levin, whereby living systems may be best understood as agents navigating abstract spaces of possible form, where mathematical structure is not merely descriptive but causally constraining. All of these frameworks push against the idea that the main explanatory unit must always be organism-level adaptation by natural selection.

Third-factor principles in linguistics (general physical principles that constrain language design independently of selection) is conceptually adjacent to natural induction. The difference is that third-factor principles are a passive constraint (physics ruling out certain structures), while natural induction is an active adaptive mechanism. Relatedly, the formalist tradition's appeal to laws of form (D'Arcy Thompson, Goodwin, Kauffman) partly overlaps with the kind of self-organization that natural induction describes.

Levin's suggestions about atemporal mathematical laws 'ingressing' into biology and offering universal constraints on forms of life is not entirely dissimilar from this biolinguistic focus on third factors (Levin 2021, 2023). The idea that there exist deep organizational motifs for organic forms strongly speaks to internalist/formalist concerns. Levin shares with researchers such as Earl Miller (e.g., Chen, Gao et al. 2026) and myself (Benítez-Burraco & Murphy 2016, 2019; Murphy 2015c, 2016a, 2016c, 2018, 2020a, 2025c, 2025d; Murphy & Benítez-Burraco 2017, 2018b) the idea that properties of bioelectric currents physically drive many core computational properties. Levin assumes some kind of latent space ('Platonic', roughly speaking), that structures in the natural world seem to draw from, and be constrained by – the universe draws from this latent space, and not the other way round. Are there laws of mathematics that seem to be thoroughly invariant to physical laws? These sorts of questions encourage Levin and others sympathetic to Platonic theories in philosophy of mathematics to assume that life has evolved through being sculpted not simply by selectional logic, but by a latent space of mathematical forms. We need not commit ourselves to a Platonic or dualistic metaphysics here, but it is worth reflecting on why it is that when many scientific questions become boiled down to their essence they essentially get exported to the mathematics department. Indeed, we also do not need to embrace any flavor of preformationism about the origin of mathematical forms constraining entities in the world, and can readily formulate a constructivist account of how third-factor constraints contribute to explanations of human language, and other processes. But this is a task for a more philosophically intense forum, so I will leave it aside here.

There remain a number of open challenges for the typical biolinguistic perspective on language evolution, and emerging findings that might force a re-location or re-contextualization of 'where' Merge-related processes can be explored. For example, recent work in evolutionary genomics reinforces a fundamentally non-saltational, mosaic view of human cognitive evolution, in which the biological substrates supporting language and other aspects of the "human condition" emerged through the gradual accumulation and recombination of genetic and developmental changes rather than a single mutation or innovation (Caporale et al. 2025). Crucially, these changes are not primarily located in protein-coding genes, but in regulatory variants shaping spatiotemporal patterns of gene expression during neurodevelopment, suggesting that human-specific cognition may reflect alterations in developmental timing, cortical organization, and circuit formation rather than entirely novel structures. This perspective shifts the explanatory target from genes directly encoding cognitive traits (e.g., language) to neurodevelopmental endophenotypes, such as extended neuronal maturation and reconfigured brain growth trajectories, which provide a mechanistic bridge between genomic variation and higher-order cognition. Within a biolinguistic framework, such findings support the view that language evolution is constrained and shaped by general biological principles (particularly developmental and regulatory dynamics) thereby aligning with third-factor approaches in which the form of linguistic systems reflects deeper properties of neural growth, timing, and organization rather than domain-specific genetic encoding alone.

As I have argued, biolinguistics should not be seen as some kind of crude 'anti-evolutionary' approach, a form of neo-Lamarckism, or a spiritual successor to early 20th-century mutationism. It is deeply concerned with the full suite of possible evolutionary causal-mechanistic forces that might have resulted in the emergence of human language. It focuses principally on the design-like properties of humans that seem to orchestrate their faculty of language, playing just as much close attention to the structure of the organism as to its place in the environment. More than other contemporary approach to language evolution, it stresses the explanatory centrality of mathematical laws that seem to guide and constrain the logic of evolution and natural selection. Biolinguistics tends to privilege explanation by internal constraints, formal organization, and third-factor principles, while not denying a role for selection. This historical tension between formalist and functionalist accounts is not merely of historiographical interest – it maps directly onto contemporary debates about what neuroscientists should predict about the brain. Adaptationist and functionalist accounts, focusing on communicative utility, tend to predict diffuse, pragmatically organized neural distributions, or they predict limited contributions from language to higher-order aspects of thought. Formalist and laws-of-form accounts predict something tighter: that the algebraic primitives of MERGE-based syntax should have specific, neurally distinguishable implementations, enforcement mechanisms that track hierarchical constituency rather than linear order, and dissociations from systems subserving non-syntactic communicative functions. Unsurprisingly, the biolinguistic approach to the brain is equally concerned with these laws of form, and the remainder of this chapter will focus on this more recently developed set of questions.

## 5. The Cortex is Not a Cave: Neural Realization and Platonic Forms

> "[I]t is the mentalistic studies that will ultimately be of greatest value for the investigation of neurophysiological mechanisms, since they alone are concerned with determining abstractly the properties that such mechanisms must exhibit and the functions they must perform".
>
> Chomsky (1965: 193)

During my time as an undergraduate student in 2011, I read the excellent volume *The Biolinguistic Enterprise*, masterfully compiled and edited by Di Sciullo and Boeckx (Di Sciullo & Boeckx 2011). Alongside the highly engaging writings of David Poeppel, which I encountered soon after (e.g., Poeppel 2012), these works were some of the primary catalysts for my interest in the biology of language. Looking back now, there was a stark lack of focus in Di Sciullo and Boeckx's volume on neurolinguistics, and perhaps too much focus on the promise of research into the genetics of language. In the intervening years, a number of advances in the cognitive neurosciences have, often indirectly, contributed to questions within biolinguistics. The question of how the formal properties of language are neurally orchestrated is still be far from being answered, but some emerging tools in the field have helped narrow the search space somewhat. In 2017, Fitch (2017: 451) complained about "how rarely formal and computational linguistic considerations play a role in current discussions of the neural basis of language" – a decade on, this complaint remains highly relevant. This section will briefly touch on some of these issues.

As Aboitiz (2017: 2) puts it, all accounts concerning language, "including genetic, cultural or linguistic accounts, will eventually have to be subordinated to an explanation of how our brains construct language". The biolinguistic enterprise is not satisfied with casual and vague descriptions of language as *a system to express meanings and communicate ideas*, or *a mode of cultural exchange*, and so forth. A MERGE-based system offers a computationally explicit means to separate out linguistic from non-linguistic forms of cognition, a tremendously important contribution to the fields of neuroscience and psychology. For instance, while usage-based models and construction grammar accounts have been highly productive at unveiling interesting psycholinguistic dimensions of syntax (Díaz-Campos & Balasch 2023), they have arguably had less success at offering possible neural mechanisms for language (answering not just *where* language is processed in the brain, but *how*). This is precisely because it is very difficult to think of ways to invite linking hypotheses when one's theories of language are not formally precise enough to adjudicate between multiple, competing neural mechanisms (an objection that in essence goes back to Chomsky 1957, 1959; Lenneberg 1967). These approaches often highlight statistical learning and gradual abstraction, pushing back against strong poverty-of-stimulus arguments and against a sharply delimited, language-specific syntactic core.

Importantly, when some researchers present linguistic word-like elements as 'cultural' products, this still presupposes some mental faculty that enters into the representation and processing of such products, and indeed, more worryingly, full-fledged combinatorial properties of morphemes (like a verb's selectional properties) are surely not 'cultural' elements of the environment (see also Krauska & Lau 2023). Some of the more interesting and novel theories in the literature concern a thoughtful integration of domain-general learning biases and principles of language design (Yang et al. 2017).

Once we take the formal properties of language seriously, as a guide for theory-formation in neighboring sciences such as neuroscience and genetics, one's scientific output and scope of explanatory models becomes remarkably more productive. This should come as no surprise to historians of science familiar with advances in chemistry and physics in the early-to-mid 20$^{th}$ century.

As outlined above, language is governed by formal constraints that fundamentally differ from other cognitive systems. Syntax instantiates a free non-associative algebra over the lexicon, but most other cognitive combinatorial systems (using vector addition, averaging, convolution, reinforcement learning value updates) are associative and structure-flattening. Therefore, neural population dynamics may exhibit specific geometric properties via algebraic operations preserving hierarchical (not just sequential) structure. As Huijbregts (2026) notes:

> "The intersection of the computational properties of biologically driven human language (i.e., a magma algebra with a merge operation) and animal call systems (i.e., a monoid algebra with concatenation as its operation) is empty. There's no overlap. Merge is commutative and non-associative while concatenation is associative and non-commutative."

Figure 2 depicts a range of neural structures, with their possible algebraic properties. Some of these may have causal power, others are non-causal mathematical models, but they all can contribute *explanatory* power to theories of neural computation (Ross & Bassett 2024). Indeed, as Gold (2006) notes, mathematical objects may be abstract, but they are not necessarily acausal, since they can be essential to, and ineliminable from, causal explanations. As discussed below, recent biolinguistics-aligned neural accounts of language processing have explicitly made use of certain of these structures, following the long tradition in neurolinguistics of trying to derive linking hypotheses between biophysics and linguistic computation. Note that the categories of structures in Figure 2 are not mutually exclusive; a single neural process may be described across multiple registers, but they do admit for distinct causal logics.

## EXPLANATORY STRUCTURES IN NEUROSCIENCE

Common metaphor • Possible formal properties • Possible linguistic relevance

### Mechanism

Hierarchical/part-whole
Fine-grained detail
Narrow (ion channel)
Broad (circuit-level)

"Clockwork"
Mereological hierarchy
Feature activation

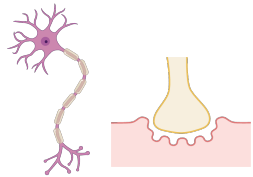

### Topology

Structural-constraint
Neural geometries
Abstracted from temporality
Manifolds

"Landscape"
Constraint space over state manifold
Semantic space

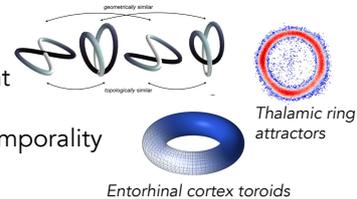

*Thalamic ring attractors*
*Entorhinal cortex toroids*

### Pathway

Sequential flow
Linear, stepwise
Dictates routes

"Highway"
Ordered monoid under concatenation
Linearization

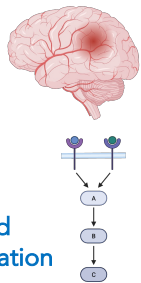

### Circuit

Fixed, closed-loop
Recurrent system
Mesoscale focus
Non-reductive
Drive oscillatory motifs

"Wiring diagram"
Fixed-point attractor
Maintenance/memory

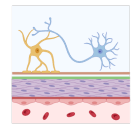

### Cascade

Initial trigger
Amplification
Momentum

"Snowball effect"
Dynamical amplification
(nonlinear differential flow)
Incremental feature assembly

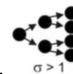

**Figure 2: Some possible explanatory structures in neuroscience**. Neuroscientific explanations can be organized into distinct formal regimes, each corresponding to different algebraic or dynamical structures. The possible formal properties attributed, hypothetically, to neural structures offer a mix of algebraic structures, dynamical systems, and ontological relations – an initial indication of where theoretical neuroscience could move towards. For further discussion of neural mechanisms, see Ross and Bassett (2024), and for possible linguistic relevance of these structures, see Murphy (2025c). Graphical elements made in BioRender (biorender.com) combined with original visualizations.

Usage-based models have not yet, to my knowledge, yielded comparably specific commitments about which formal classes of neural mechanism (as in Figure 2) should implement syntactic computation. On the other hand, once we have a clear algebraic account of syntax as being a free, non-associative commutative magma MERGE-based operator, we can narrow down target neural structures accordingly. To echo one of Chomsky's regular lines, this is simply standard scientific practice: taking a computational-level account and using it to guide exploration in other domains.

Of course, I should highlight here that the biolinguistic approach may be entirely inappropriate for exploring how the human brain processes language, in the sense that there is never any necessary connection between our mathematical models and the behavior of neural tissue. But, to my knowledge, there are no stronger candidates from the 'bottom-up' side of the field, mostly because a substantial portion of work in the neuroscience of language treats language as roughly equivalent to *lexical prediction*, tying it closely to statistical features such as surprisal and entropy.

One of the major issues of using surprisal to measure neural responses to language is that surprisal is too 'over-powered' and all-encompassing, and captures statistical effects of language *in addition* to a host of non-linguistic features (Slaats & Martin 2025). Surprisal is a very strong descriptor of behavioral and neural data precisely because it is representation-agnostic and can absorb variance from many latent sources at once. That makes it useful as a predictor, but limited as an explanation of mechanism. Slaats and Martin (2025) argue that surprisal is an excellent predictor of behavioral and neural responses not because it transparently reveals the mechanisms of language comprehension, but because it is a representation-agnostic probabilistic summary that pools over multiple latent sources of variance, including syntax, lexical frequency, and contextual regularities. As such, surprisal may track consequences of structure without constituting structure or explaining how structured meaning is computed. This is especially important for LLM-brain alignment work: if an LLM-derived surprisal signal aligns with neural data, that alignment does not show that the model has recovered the brain's mechanisms for building hierarchical syntax or computing compositional meaning. Rather, it may only show that both brains and models are sensitive to broad statistical regularities in sequential input. As usual, statistical information remains a useful *cue* to structure, but is not a substitute for it.

The relevant contrast, then, is not between statistical and non-statistical systems in any absolute sense. Neural systems will of course be sensitive to statistical regularities. The critical issue is whether the mechanisms in question preserve relational asymmetries and type-sensitive compositional structure, or whether they merely compress co-occurrence patterns. The biolinguistic wager is that syntax requires the former, and that this requirement should be detectable in the causal organization of neural dynamics.

Effectively moving against some contemporary 'inside-out' accounts of cognitive neuroscience (e.g., Buzsáki 2019), the biolinguistics approach typically assumes that "no notion of competence will emerge from simply looking at the brain" (Mukherji 2022: 69). Unsurprisingly, if we define language as heavily in the game of predictive processing, then we will find many neural indicators of this, since the brain is certainly, for the most part (but not entirely), a predictive organ. The Bayesian brain hypothesis has rightly had many successes in domains of cognitive neuroscience, but we ultimately must reconcile with the fact that language is *not* at its core about prediction (for a recent critique that expands on this point, see Murphy 2025a). Indeed, "[t]he 'essence' of language, if you will, is mathematical" (Watumull 2013: 306). Still, it is certainly easier to ignore these formal and algebraic constraints, and to treat language as something vague like "sentence-level meaning", which I think explains the sheer number of researchers who stick to simpler formulations of language when investigating its neural correlates. We might ask, have we learned any new facts about language from the recent decade or so of brain-LLM alignment research, treating language as a predictive organ? It is not immediately obvious that we have. There have been a few attempts to migrate the concerns of theoretical linguistics to other domains of the cognitive sciences (see Murphy et al. 2024a), but neuroscientific models of language that offer a clear definition of what *language* is meant to be are surprisingly uncommon. Meanwhile, many neuroscientists remains highly dogmatic about the assumption that all forms of human behavior will ultimately yield to the same level of description (e.g., basic cell assemblies), a bias that many other fields shook off decades ago before embracing notions from complexity science (Krakauer 2024).

The reason why biolinguistic interventions are critically important in the neurosciences is because natural language syntax is not remotely similar to other cognitive faculties like music and mathematics, despite some superficial similarities. MERGE is inherently Markovian; it cannot "see" and has no memory of what came before it, it only applies itself to a given workspace with no extensive lookback memory into past applications. MERGE is Markovian in the algebraic sense, though this is not to be confused with Markov models in the sequential-statistical sense commonly invoked in computational neuroscience. It produces an obligatory symmetry-breaking labeling algorithm that always categorizes syntactic structures (Chomsky 2013, 2014, 2015), whereas music and math tolerate symmetry with little complaint. The history of the derivation is not preserved in the current stage, and 'Minimal Yield' makes derivations strictly Markovian in that the next step has no access to derivational history. This, critically, renders linguistic recursion distinct from other types of recursion. In addition, recursion is principally distinct from notions like self-embedding, and classic rewriting rules return strings, not structures, hence some recent focus on the algebraic properties of MERGE in order to isolate the simplest properties of language (Berwick & Chomsky 2019) (*what* a recursive operation does and *how* it proceeds are not the same thing). It is unclear how contemporary approaches within cognitive neuroscience that centre around brain-LLM alignments would deal with these domain-specific restrictions on application.

At the same time, others have rightly cautioned that this focus on formal properties of language might lead us down unhelpful rabbit holes, depending on our methods. For example, Mukherji (2022) objects to theories within the biolinguistic framework that are based on the idea that because minimalism uses binary branching in much of its formalism, these operations can thereby be rewritten in the binary notation vector of physics (e.g., Piattelli-Palmarini & Vitiello 2015). Steps of syntactic derivation could then be called vector states, but nothing in terms of explanation is gained by comparing these remote formal systems. Mukherji is quick to call out these and other similar proposals that purport some force towards explanatory adequacy yet ultimately appear an awful lot like re-description.

Synthesizing these ideas with some of the previous themes discussed in this chapter, we can see that work within the biolinguistic tradition is sympathetic to moves within mechanistic philosophy of science (Craver 2007; Machamer et al. 2000), summarized by Craver and Darden (2013: 15):

> "Mechanisms are how things work, and in learning how things work we learn ways to do work with them. Biologists try to discover mechanisms because mechanisms are important for prediction, explanation, and control. […] Biologists seek mechanisms that produce, underlie, or maintain a phenomenon".

With these important qualifications noted, some recent efforts have been made to reconcile the neural code for hierarchical syntax with predictive processes. For instance, Ramírez (2015) offers a specific oscillatory metaphor for locality effects in syntax. Baggio (2018) discusses some formal properties of semantics that might align with the processing of certain brain regions. A line of research (starting with Ding et al. 2016) has revealed how rhythmic neural activity can be organized to track hierarchical aspects of language from speech (Benítez-Burraco et al. 2023; van Bree 2024).

More recently, the ROSE model offers a neurocomputational account of how syntax is neurally processed (Murphy 2024, 2025c). Under this model, instructions for symbolic phrase structure representations interface with probabilistic aspects of linguistic processing, with different types of cross-frequency coupling being hypothesized to interface these domains. A frontotemporal *symbolic low-frequency phase code* interacts via cross-frequency coupling with a series of local *probabilistic inferences over lexico-semantic content*, with the latter being implemented via spike-phase coupling assembling bundles of linguistic features. This low-frequency phase code actively gates and modulates local spiking assemblies that represent lexical items. ROSE tries to give a neural implementation of a syntax whose formal properties are non-associative, commutative, hierarchical, and nonplanar.

ROSE is built on the premise that a core (and deeply puzzling) component of language is its compliance with *structure-dependence* over linear rules: "[W]e ignore the simple computation on linear order of words [adjacency], and reflexively carry out a computation on abstract structure" (Chomsky 2022). The model provides a possible infrastructure for flexibly implementing distinct types of minimalist grammar parsers for the real-time processing of language. This perspective helps to furnish a more restrictive 'core language network' in the brain than approaches that isolate general sentence composition: the language network is defined in Murphy (2025c) as being causally involved in executing specific parsing operations (i.e., establishing phrasal categories, tree-structure depth, resolving dependencies, and retrieving proprietary lexico-syntactic representations), capturing these network-defining operations jointly with probabilistic aspects of parsing. ROSE can be seen as offering a 'mesoscopic protectorate' for natural language (Laughlin et al. 2000), whereby syntax is a complex emergent phenomenon that cannot be approached sufficiently by looking under-the-hood purely at the lower-level 'mechanistic' levels of neurobiological organization (Figure 2). ROSE can also be leveraged to contribute to explanations of syntactic transfer in L2 sentence production (Uluslu & Murphy 2026).

How do neural models like ROSE relate to the explanatory structures in Figure 2? Cortical cascades have been shown to support rapid semantic inferences during reading (Murphy et al. 2026), assembling relevant features of words. Meanwhile, ROSE makes use of certain circuit-level mechanisms and topological structures. ROSE treats syntax as implemented by nested causal processes spanning local clusters, cross-frequency coupling, and inter-areal traveling waves. It also appeals to recurrent local loops, workspace maintenance, and dynamical motifs as candidate substrates for stable feature-bundling and parsing states, thereby appealing also to the circuit register (Figure 2). Dynamical motifs are relevant to the R/O levels, but topology is not yet the core explanatory language of ROSE: In the current formulation, topological structures act more like an interface-level description than the central implementational story. ROSE mainly explains syntax through oscillatory coordination. Topology enters where one wants to characterize the shape of conceptual spaces, state manifolds, or complexity classes of neural computation.

Though many speculative hypotheses remain to be further tested, the main theme here is that if syntax is defined by algebraic/topological constraints, and not by sequential concatenation, then this may provide guidance for possible neural explanatory structures (in Figure 2). If the brain implements MERGE, we should expect syntactic representations to occupy a specifically shaped region of neural state space, one where the geometry encodes constituency and hierarchical depth, not just semantic similarity.

Neurocomputational models such as this are often proposed within the spirit of neurosymbolic approaches that jointly leverage the strengths of both symbolic and connectionist models. As Fitch (2009: 296) notes, "a connectionist model at the implementational or algorithmic level is not necessarily in conflict with a symbolic computational model, but rather a potential complement to it". For instance, ROSE is assembled from a constellation of generic neurocomputational procedures, in the spirit of seeking combinations of conserved cell processes building neural circuits that end up performing qualitatively unique classes of computation (Szathmáry 2001).

As is often the case, some other formal considerations can help narrow the hypothesis space for neural implementation. If human sentence processing, in its full expressive power, occupies the mildly-context sensitive level of the formal language hierarchy (Stabler 2004), this additional memory involved in parsing grammatical structures beyond the finite-state level should have the characteristics of a queue, not a stack (Fitch 2009). Some recent hypotheses have been offered concerning the locus of syntactic buffers, within a tripartite syntax network involving lateral posterior superior temporal sulcus, posterior inferior frontal cortex, and posterior middle temporal gyrus (Murphy 2026). Further work is needed to explore the dynamic interactions between these sites during the parsing of syntactically complex expressions.

In summary, a formal theory of syntax should constrain what counts as an adequate neural theory, rather than allowing neuroscience to drift into vague notions like "sentence meaning" or generic predictive processing.

## 6. Assembling the Pieces

These ideas provide additional hints at evolutionary plausibility for the biolinguistic approach. How might be synthesize some of the ideas presented in this chapter, with an eye towards explanatory accounts of evolvability, learnability, and neural orchestration?

Syntax may have emerged when evolution discovered a way for neural tissue to convert transient multimodal feature coalitions into recursively re-usable, type-stable labeled objects, involving minor modification to multiple of the elements in Figure 2, but in ways that, when assembled together, yielded a marked shift in computational resources. To convert this into dynamical terms (which may, or may not, end up being the most suitable terminology), when two or more feature complexes were brought into a shared workspace, the system could settle into a new attractor state that was not merely a mixture of the parts, but a new categorized object with a dominant type identity. This was not merely an increase in memory, an increase in information capacity (as per Cantlon & Piantadosi 2024), or sequencing ability, or representational richness, but a qualitative transformation in the format of cognition itself. On this view, the evolutionary origin of syntax lies not in the gradual accumulation of combinatorial operations (i.e., some form of proto-language evolving into something more complex), but in the emergence of a system capable of treating combinations as *new kinds of things*. Once such a system exists, recursion follows as a structural consequence: the output of one operation is already in the correct format to serve as the input to another.

If syntax emerged as a phase transition in neural organization rather than through gradual combinatorial accretion, we might expect several empirical signatures. First, the neural mechanisms supporting hierarchical composition should be qualitatively distinct from those supporting sequential concatenation — not merely more of the same, but formally different in kind. Emerging evidence is reviewed in Murphy (2025c). Second, we might expect the syntactic capacity to be relatively encapsulated developmentally, appearing rapidly once a critical neural configuration is in place, rather than building up incrementally. The speed and uniformity of syntactic acquisition across languages and cultures is at least consistent with this prediction. Third, we should expect that damage to the relevant neural configuration produces catastrophic rather than graded syntactic deficits – a prediction that receives some support from lesion studies showing sharp dissociations between syntactic and other linguistic capacities (Matchin et al. 2022; Matchin & Hickok 2020).

How might we relate these ideas to the above discussion of natural induction? Assume that evolving nervous systems were repeatedly perturbed by pressures from planning, social cognition, tool use, vocal control, and conceptual integration. These pressures did not directly select syntax piece by piece; rather, they pushed cortical systems into a region of state space where stable labeled closure (meaning, the closure of MERGE-based objects) became developmentally learnable and neurally reproducible. Once found, this regime was so computationally powerful that selection canalized it. In that sense, syntax is not a hill-climbed adaptation in the usual picture, but rather it is a discovered phase of neural organization. This perspective preserves a saltationist origin of syntax while dissolving the apparent explanatory gap: selection does not construct syntax, but rather inherits and stabilizes a computational phase first discovered through the intrinsic dynamics of neural organization.

The biological neural networks that implement language processing are dynamical systems with viscoelastic-like connection properties; synaptic plasticity being the biological analog of the spring-like connections in Watson et al.'s (2025) formalism. These networks are subjected to environmental perturbation (experience, input) and their connection weights adjust slowly over both developmental and possibly evolutionary timescales. The result, on Watson et al.'s account, would be the *spontaneous emergence of adaptive organization* in the form of associative generalizations that are more than merely locally optimal, which is a plausible description of how hierarchical syntactic structure might emerge as a generalizing solution to the problem of assembling structured thought. The 'induction-first' version of phenotype-first evolution (Watson et al. 2025) offers an account of how within-lifetime neural adaptation can guide subsequent genetic evolution: the neural network finds the adaptive configuration first, and genetic change consolidates it. This is consistent with the biolinguistic emphasis on laws of form and physical constraints: the mathematical properties of hierarchical structure make it the natural attractor for associative neural systems, which then show natural selection what to do.

However, there is some initial tension between natural induction's associativity and the formal nature of MERGE's non-associativity. If natural induction is essentially a sophisticated associative/learning process, and if biolinguistics insists that hierarchical syntactic structure requires something *beyond* associative generalization (because structured relations cannot be reduced to similarity-based associations), then natural induction as described may not reach all the way to the kind of structural organization that MERGE delivers. Natural induction might explain the emergence of certain associative patterns in neural organization, but it is not obvious that it can generate the specific algebraic properties of MERGE. Natural induction might explain why neural systems converge on *some* form of hierarchical chunking, without that chunking having the full algebraic properties of MERGE. The specific formal properties of MERGE might then require additional explanation, either through third-factor mathematical constraints or through a minimal genetic specification.

In short, Watson et al.'s (2025) mechanism is entirely appropriate for explaining how neural networks learn to associate phonological forms with conceptual content, how children rapidly generalize lexical categories from sparse input, or how the externalisation component adapts to particular language-specific linear orders, in addition to other types of 'interface' relations. These are all associative tasks, and natural induction provides a principled non-selectionist account of how they get learned and stabilized.

Turning to future directions for the topics in neurobiology discussed here, consider the work of Marcolli and Berwick (2026). These authors draw principled connections between neural signatures like phase coherence and the category-theoretic formulation of MERGE, hence tentatively suggesting, for the first time, causal links between computational models and biological implementation. Marcolli and Berwick (2026) show that if lexical items can be encoded as functions in a wave-like function space, then arbitrary syntactic objects and MERGE operations can also be embedded there in a mathematically faithful way, preserving the crucial algebraic properties of free symmetric Merge (especially commutativity and non-associativity) by means of a non-associative commutative semiring built using second Rényi entropy. They then show how this abstract construction can be realized, at least as a proof of concept, through neural-style circuits and, in a simplified case, through phase synchronization and cross-frequency coupling on sinusoidal waves. This work is particularly important because it does not merely gesture at oscillations as a possible implementation language for syntax, but offers what may be the clearest formal bridge yet between category-theoretic/algebraic models of MERGE and neurocomputational mechanisms such as phase synchronization, phase-amplitude coupling, and related phenomena (Yang et al. 2025). For further discussion of how these neural processes can enforce the algebraic properties of language, see Murphy (2025c). These works are sympathetic to Turing's belied that formal theories could offer novel "restrictions" on theories of brain function (Coveney & Highfield 1995: 388).

In developing these kinds of neural models of language, and working within mathematical biolinguistics more generally (which has traditionally defined I-language as a Turing machine), we are not to be accused of falsely equating the abstract with the physical, but rather we have "abstracted away from the contingencies of the physical, and thereby discovered the mathematical constants that […] must of necessity be implemented for any system — here biological — to be linguistic. This abstraction from the physical is part and parcel of the methodology and, more importantly, *the metaphysics* of normal science" (Watumull 2013: 310). Indeed, these forms of idealizations are closest things we have to approximating the world as it truly is (Watumull 2013: 310–311):

> "Consider Euclidean objects (e.g., dimensionless points, breadthless lines, perfect circles, and the like). These objects do not exist in the physical world. The points, lines,

> and circles drawn by geometers are but imperfect approximations of abstract Forms
> — the objects *in themselves* — which constitute the ontology of geometry".

As was understood by Plato, physical reality is an inconstant, and deceptive, surface, underneath which exists the perfect structures of nature, intrinsically formal in character. The reason why the cognitive neurosciences should not forget this is because *all* of the greatest success stories in the natural sciences have had to learn this the hard way.

Contrary to many misconceptions, the notion of language as a system of discrete infinity (via some type of MERGE operator) is not incompatible or inconsistent with the assumption that language has a finite, biological basis. The consilience is effected by Turing's proof of the coherency of a finitary procedure (i.e., defining the computable numbers as those determinable by his mathematical machine) generative of infinite sets (Watumull 2013). Many neuroscientists in the 2020s have limited their operational definition of language to the sorts of algorithms and structures that are safely 'neurobiologically plausible', without much regard to whether or not those structures (often predictive, sequential) bear any principled relation to the behavior under investigation (for excellent discussion of related themes, see Krakauer et al. 2017). Moreover, we plainly do no know enough about the physical and computational properties of neurons to make any strong preliminary conclusions that the formal properties of MERGE are incompatible with them (Murphy 2025b). This has long been recognized in the biolinguistics tradition, most notably by Chomsky (2004b: 41–42):

> "[E]ven though we have a finite brain, that brain is really more like the control unit for an infinite computer. That is, a finite automaton is limited strictly to its own memory capacity, and we are not. We are like a Turing machine in the sense that although we have a finite control unit for a brain, nevertheless we can use indefinite amounts of memory that are given to us externally to perform more and more complex computations. […] We do not have to learn anything new to extend our capacities in this way".

To conclude, the biolinguistic view of language is sympathetic to the way Wilhelm von Humboldt construed it. He stressed how language is not a product (*Ergon*) but an activity (*Energeia*) (Underhill 2009: 58), or the infinite use of finite means, to invoke the rationalist-romantic conception. The importance of biologically grounded, neurocomputational approach to the mind should not be overlooked, with Hassabis (2012: 462) suggesting that "from a neuroscience perspective, attempting to distil intelligence into an algorithmic construct may prove to be the best path to understanding some of the enduring mysteries of our minds". The biolinguistic tradition seeks to continue this mission.

## 7. Conclusion

To return to this chapter's central contention: biolinguistics still matters because it insists that the object of inquiry be defined with sufficient formal precision to generate testable neural predictions. Where other approaches treat language as a diffuse capacity for meaning integration or statistical prediction, recent algebraic characterizations of MERGE yield specific expectations about neural structure and geometry – expectations that neural models are now beginning to probe. The field stands at a stark inflection point: category-theoretic formalizations of syntactic structure are, for the first time, being connected to concrete neural mechanisms via principled mathematical bridges, not metaphors. Whether these bridges hold under the weight of further empirical scrutiny will determine much about the future of the program. But the deeper lesson of the biolinguistic tradition is not tied to the fate of any single model. It is that the study of language, like the study of light or of chemical bonds before it, advances most decisively when we stop cataloguing surface regularities and begin asking what formal structures nature must implement to produce the phenomena we observe. As mentioned, Plato urged that we proceed by means of problems and leave the starry heavens alone. The biolinguistic enterprise has taken this counsel seriously.